\documentclass[letterpaper, 10 pt, conference]{ieeeconf}  % Comment this line out if you need a4paper

\IEEEoverridecommandlockouts                              % This command is only needed if 
\usepackage{amssymb}
\usepackage{amsmath}
\usepackage{booktabs}
\usepackage{graphicx}
\usepackage{multirow}
\usepackage{bigdelim}
\usepackage{pifont}
\usepackage{bbding}
\usepackage{titlesec}
\usepackage{algorithm}
\usepackage{algorithmic}
\usepackage{tabularx}
\title{\LARGE \bf
RPG: Robust Policy Gating for Smooth Multi-Skill Transitions in Humanoid Fighting
}

\author{%
  Yucheng Xin$^{1,2}$, Jiacheng Bao$^1$, Yubo Dong$^2$, Dong Wang$^{2,3}$, Junbo Tan$^2$, Xueqian Wang$^2$, Bin Zhao$^{\dagger1}$, Xuelong Li$^{1,2}$\\
}

\author{%
}
\author{Yucheng Xin$^{1,2*}$, Jiacheng Bao$^{2*}$, Yubo Dong$^{3,2}$, Xueqian Wang$^{1}$, Bin Zhao$^{2}$\\
Xuelong Li$^{2}$, Junbo Tan$^{1\dagger}$, Dong Wang$^{2\dagger}$% <-this % stops a space
\thanks{$^{1}$Center for Artificial Intelligence and Robotics, Shenzhen International Graduate School, Tsinghua University, China, {\tt\small \{xin-yc23@mails., wang.xq@sz., tjblql@sz.\}tsinghua.edu.cn}, $^{2}$Shanghai AI Laboratory, $^{3}$Shanghai Jiao Tong University}
\thanks{$^\dagger$Corresponding author, $^{*}$indicates equal contribution}
}

\begin{document}

\maketitle
\thispagestyle{empty}
\pagestyle{empty}

%%%%%%%%%%%%%%%%%%%%%%%%%%%%%%%%%%%%%%%%%%%%%%%%%%%%%%%%%%%%%%%%%%%%%%%%%%%%%%%%

\begin{figure*}[ht]
  \centering
  \includegraphics[width=\linewidth]{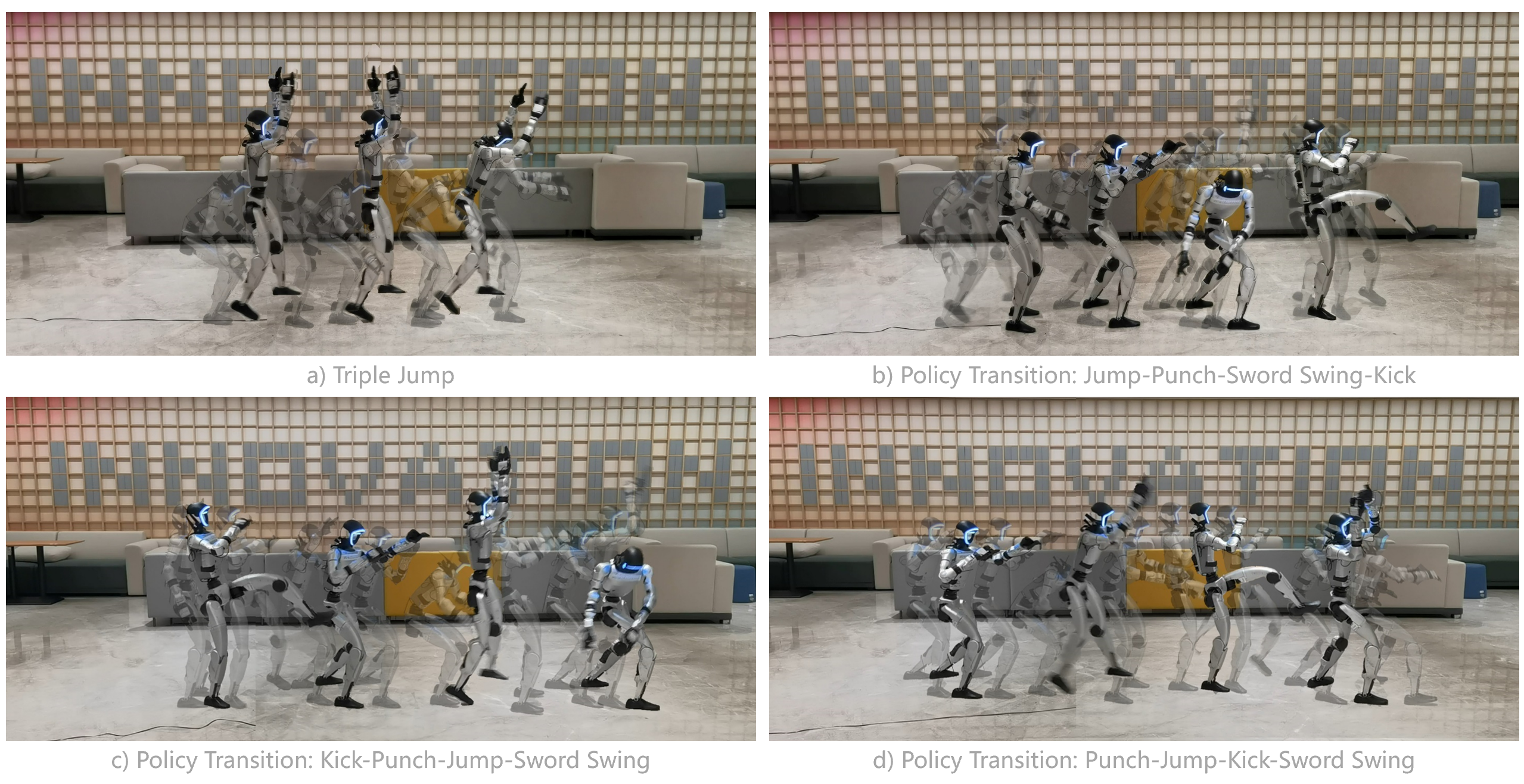}
  \caption{\textbf{Policy Transition Demonstration.} We conducted policy transition tests for robotic combat motions using the proposed RPG. Punching and Sword Swing motions primarily involve the upper body, whereas Jumping and Kicking motions are mainly lower-body actions. We demonstrate 4 distinct policy transition combinations here, highlighting the motion capabilities during transition between upper and lower-body strategies.
  \textbf{a)} Jumping introduces significant instability for humanoid robot. However, by repeatedly executing the jumping policy in succession, we achieved a triple jump sequence.
  \textbf{b)-d)} Other policy transition combinations.
  }
  \label{fig:motion}
\end{figure*}

\begin{figure*}[ht]
  \centering
  \includegraphics[width=\linewidth]{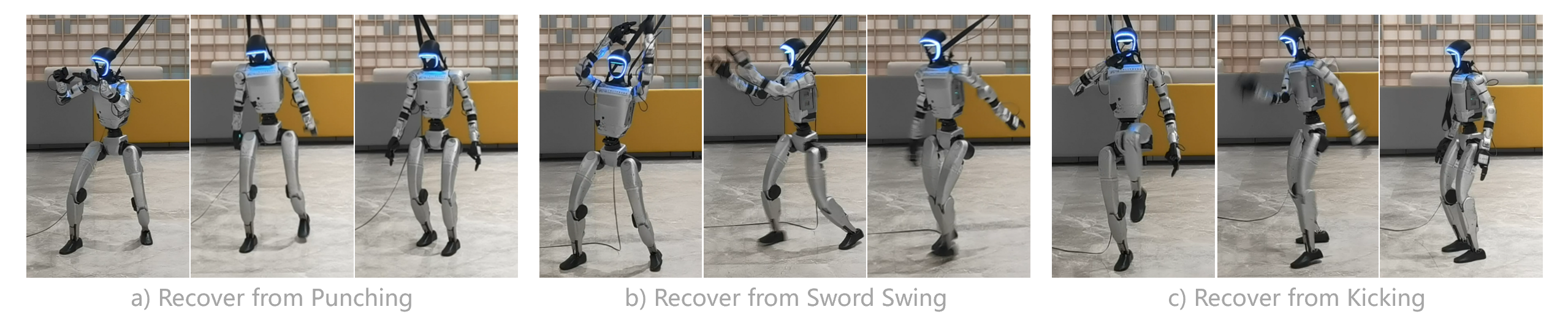}
  \caption{\textbf{Recovery Tests.} Due to the short duration of the Jumping motion, it was excluded from consideration. For the other motions, the robot demonstrated the ability to avoid foot sliding and resume a stable, human-like standing posture from any interrupted state during execution.}
  \label{fig:recover}
\end{figure*}

\begin{abstract}

Humanoid robots have demonstrated impressive motor skills in a wide range of tasks, yet whole-body control for humanlike long-time, dynamic fighting remains particularly challenging due to the stringent requirements on agility and stability. While imitation learning enables robots to execute human-like fighting skills, existing approaches often rely on switching among multiple single-skill policies or employing a general policy to imitate input reference motions. These strategies suffer from instability when transitioning between skills, as the mismatch of initial and terminal states across skills or reference motions introduces out-of-domain disturbances, resulting in unsmooth or unstable behaviors. In this work, we propose RPG, a hybrid expert policy framework, for smooth and stable humanoid multi-skills transition. Our approach incorporates motion transition randomization and temporal randomization to train a unified policy that generates agile fighting actions with stability and smoothness during skill transitions. Furthermore, we design a control pipeline that integrates walking/running locomotion with fighting skills, allowing humanlike long-time combat of arbitrary duration that can be seamlessly interrupted or transit action policies at any time. Extensive experiments in simulation demonstrate the effectiveness of the proposed framework, and real-world deployment on the Unitree G1 humanoid robot further validates its robustness and applicability.

\end{abstract}

\section{INTRODUCTION}

Humanoid robots have shown remarkable progress in recent years, demonstrating agile 
% locomotion~\cite{he2025attentionbaseda, radosavovic2024learning, zhuang2024humanoid, wang2025dribble, xue2025unified, lin2025let, zhang2025learning},  
locomotion~\cite{radosavovic2024learning, zhuang2024humanoid, wang2025dribble, xue2025unified, zhang2025learning}, 
% dexterous manipulation~\cite{ben2025homie, gu2025humanoid, liu2025unleashing, schakkal2025hierarchical, sun2025ulc, tessler2025maskedmanipulator, zhang2025falcon, dao2023simtoreal, agravante2019human} 
dexterous manipulation~\cite{ben2025homie, gu2025humanoid, liu2025unleashing, schakkal2025hierarchical, sun2025ulc, tessler2025maskedmanipulator, agravante2019human}
, and even athletic motions~\cite{he2025asap, truong2025beyondmimic, xie2025kungfubot}. A complex, practical, and compelling testbed task is enabling humanoid robots to perform human-like fighting actions such as jumping~\cite{xue2025unified, truong2025beyondmimic, zhuang2024humanoid}, punching~\cite{he2024omnih2o}, sword swings, and kicking~\cite{xie2025kungfubot, zhang2025hub}. Such task requires whole-body coordination, rapid skill switching, and stable contact control, which place much higher demands on robustness and agility compared to conventional locomotion or manipulation. An ideal humanoid whole-body control policy for fighting would resemble controlling a character in a role-playing game (RPG), where a user can seamlessly trigger and execute diverse combat skills in real time. Achieving this level of flexible and fluid multi-skill control in physical humanoid robots, however, remains a significant challenge.

Recent advances in imitation learning have enabled robots to learn complex motion skills from reference motion trajectories. While effective for single-skill learning~\cite{he2025asap, xie2025kungfubot}, extending imitation learning to multiple diverse skills is non-trivial. Naive solutions, such as switching between separate policies or using a general policy to mimic diverse reference motions~\cite{he2024omnih2o, he2024learning,chen2025gmt,zhang2025hub,su2025hitter}, often lead to instability in policy deployment. This is primarily due to mismatched initial and terminal states between reference sequences, leading to out-of-domain disturbances during skill transitions. As a result, robots suffer from jerky or unsmooth motions, particularly when attempting to concatenate different fighting skills for practical human-like combat.

To overcome these challenges, we propose Robust Policy Gating (RPG), a hybrid expert policy framework designed for multi-skill imitation learning with smooth skill transitions. Our method first trains multiple expert policies on distinct categories of fighting skills. During training, we introduce policy-transition randomization and temporal randomization to explicitly simulate mid-sequence truncations and arbitrary skill switches, forcing each expert to learn robustness against discontinuous motions. Once the experts converge, we freeze them and train a lightweight gating network that outputs weighted combinations of their actions, regularized with torque and contact smoothness objectives. The resulting hybrid expert policy controller enables fluid switching across skills, even under abrupt transitions.

Extensive experiments in simulation demonstrate that RPG effectively generates agile and seamless fighting behaviors, even under abrupt action switches. Beyond isolated fighting skills, we design a control pipeline that integrates locomotion and combat behaviors. When no fighting action is triggered, the robot maintains a locomotion state; when a skill is commanded, the system seamlessly transitions to the specified action, supporting arbitrary interruptions and varying durations. This design provides a game-like interface for controlling humanoid robots, akin to RPG game combat mechanics. Furthermore, we validate the approach on the Unitree G1 humanoid robot, confirming that RPG transfers successfully to real hardware and enables robust execution of high-agility fighting motions.

In conclusion, our contributions are as follows:
\begin{itemize}
    \item [1.] We introduce Robust Policy Gating (RPG), the framework enabling smooth and robust multi-action transitions for humanoid fighting behaviors.
    \item [2.] We introduced a novel method incorporating both policy-transition and temporal randomization during expert policies training, and a new gating network for smoothness regularization, achieving robustness against disturbances during policy transitions and stable fusion of multiple expert policies.
    \item [3.] We designed a pipeline that integrates locomotion and fighting skills, allowing users to control the robot for prolonged combat in a manner similar to playing an action RPG game.
\end{itemize}
\section{RELATED WORKS}

\subsection{Imitation Learning for Humanoid Whole-Body Control.}
Recent advances in imitation learning have empowered humanoid robots to acquire dynamic and naturalistic skills by tracking and synthesizing complex human motions. For agile behaviors, works such as \cite{he2025asap, li2023robust} focus on jumping, while PBHC \cite{xie2025kungfubot} demonstrates multi-step motion reproduction for Kungfu and dancing. BeyondMimic \cite{truong2025beyondmimic} further introduces a guided diffusion framework that enables humanoids to execute highly challenging motions, including spins, sprinting, and cartwheels, while also supporting zero-shot task-specific control.

Expressive and resilient skills have also been explored: Exbody \cite{cheng2024expressive} and Exbody2 \cite{ji2024exbody2} focus on dance imitation, whereas Embrace Collisions \cite{zhuang2025embrace} tackles recovery behaviors. Humanoid parkour has been showcased in \cite{zhuang2024humanoid, zhuang2023robot}. For contextual imitation, Videomimic \cite{allshire2025visual} leverages everyday human videos with environment reconstruction to teach humanoids real-world tasks such as stair climbing, sitting, and standing.

On scalable and general motion tracking, GMT \cite{chen2025gmt} employs adaptive sampling and a motion mixture-of-experts (MoE) to capture diverse skills with a single controller, while BumbleBee \cite{wang2025experts} distills clustered experts into a generalist policy for whole-body humanoid control. HuB \cite{zhang2025hub} further integrates motion refinement, balance-aware learning, and robustness training to master extreme balancing poses such as the Swallow Balance and Bruce Lee’s Kick.

Beyond full-body imitation, recent efforts extend to manipulation and sports. OKAMI \cite{li2024okami} and HumanPlus \cite{fu2024humanplus} emphasize upper-body skills for object interaction, and HITTER \cite{su2025hitter} combines model-based planning with reinforcement learning to achieve agile, stable table-tennis rallies.

\subsection{Policy Switching and Motion Transition.}
Enabling diverse robot skills often involves skill composition or switching mechanisms, such as R2S2’s composable skill library \cite{liu2025unleashing} or automatic gait discovery \cite{yu2025discovery}. ASE~\cite{peng2022ase} learns a continuous latent skill space through adversarial imitation, enabling smooth skill transitions without predefined segmentation. Other methods use contextual expert switching: \cite{neggatu2025evaluationtime} uses uncertainty to switch between RL and BC. For robustness, \cite{panda2025robust} formalizes policy selection as a multi-armed bandit problem.

Generating smooth and stable transitions between motions is critical for robust robotic performance and has been addressed through various methods. These include dedicated transition policies like the Expert Composer \cite{christmann2024expert} for skill sequencing, as well as techniques that engineer smoothness directly into controllers through regularization \cite{huang2025learning}, command interpolation \cite{sun2025ulc}, and comfort-oriented rewards \cite{wang2025endtoend}. Our Robust Policy Gating (RPG) framework builds on this line of work by achieving seamless transitions between diverse skills through a specific randomization method and a learned gating network.
\section{METHODS}

\begin{figure*}[ht]
  \centering
  \includegraphics[width=\linewidth]{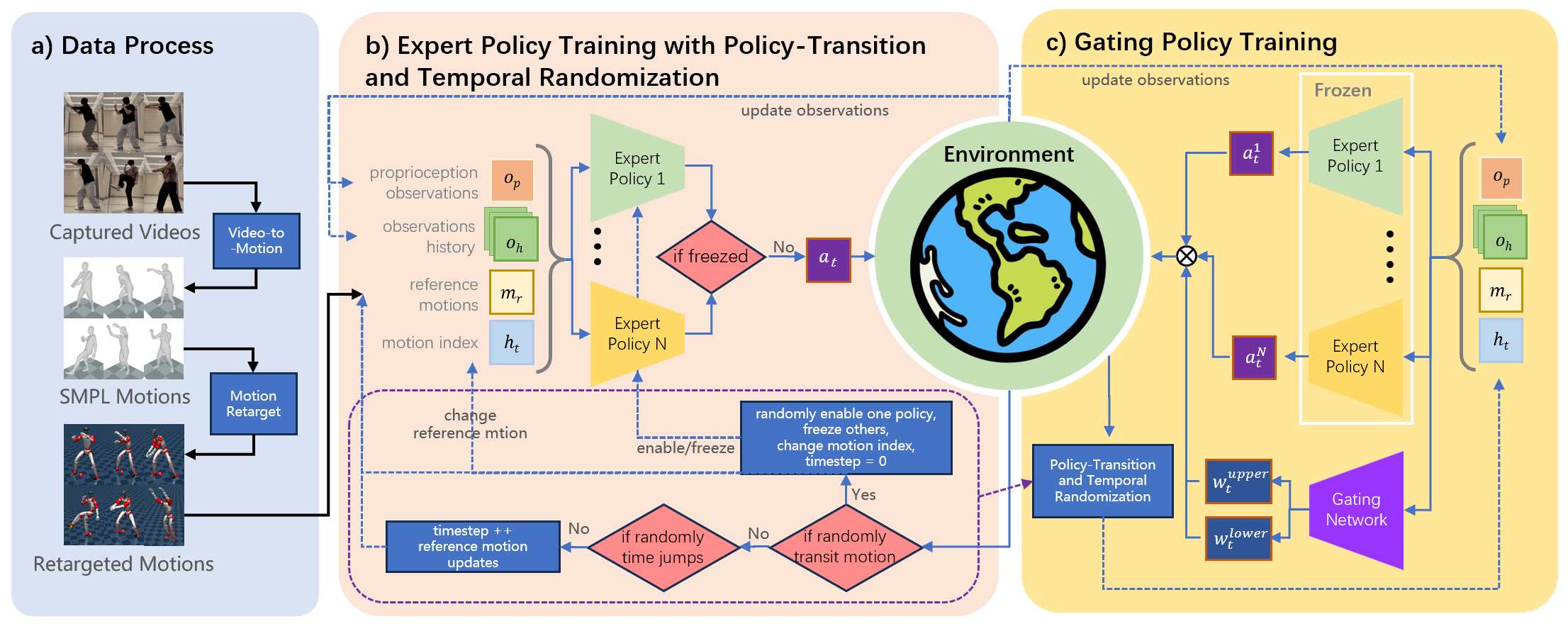}
  \caption{\textbf{Framework.} \textbf{a)} Collection of human demonstration data for combat motions (jumping, punching, sword swing, kicking), with video-based motion conversion and retargeting to adapt to the Unitree G1 humanoid robot;
  \textbf{b)} Expert networks are trained for each motion type via imitation learning. All experts are loaded concurrently, though only one policy is updated per step. Policy-transition and temporal randomization are introduced: at any time step, the system may stochastically switch to a new motion, freezing the current policy, resetting the reference motion, and enabling a new policy to update. Additionally, random time-jumping in reference motion sequences is applied to simulate abrupt motion changes and enhance robustness.
  \textbf{c)} After the expert policies convergence, the expert policies are frozen while the randomization method remains active. A gating network is trained to blend outputs from the experts. The action output is decomposed into components for upper and lower bodies, and the gating network computes weighted combinations for each. The gating network is optimized with smoothness regularization to improve motion fluency during policy transition.}
  \label{fig:framework}
\end{figure*}

\subsection{Framework.}
Our proposed Robust Policy Gating (RPG) framework is designed to learn a unified control policy that seamlessly executes and transitions between multiple dynamic fighting skills. The methodology consists of four core components: (i) data acquisition and motion retargeting, (ii) robust expert policies training via policy-transition and temporal randomization, (iii) a gating network for smooth policy fusion, (iv) sim-to-sim and sim-to-real validation and integration with a locomotion policy for a complete control pipeline. The overall framework is depicted in Fig. \ref{fig:framework}.

\subsection{Data Process.}
We begin by collecting motion references from two sources: (i) existing public motion datasets~\cite{mahmood2019amass}, and (ii) video recordings obtained from demonstrations or online footage. To convert raw video into motion sequences, we employ the GVHMR~\cite{shen2024worldgrounded} framework, which extracts 3D human motions from monocular video. The resulting motion clips are then processed using the PHC retargeting method~\cite{luo2023perpetual}, adapting the human motions to the humanoid robot’s kinematic structure while preserving key dynamics.
From this pipeline, we obtain a set of representative actions relevant to humanoid fighting: punching, jumping, sword swings, and kicking. These action segments form the basis for policy training, generating kinematically feasible reference motion sequences ${\mathcal{M}^{r}_m,m\in M}$, where $M=\{j,p,s,k\}$ represents motions of jumping, punching, sword swing and kicking.

% \subsection{Expert Policy Training with Policy-Transition and Temporal Randomization.}
\subsection{Expert Policy Training with Randomization Strategies.}

For each action category, we train a dedicated policy $\pi^m, m\in M$ using Proximal Policy Optimization (PPO). Each policy is optimized to imitate its corresponding reference motion $\mathcal{M}^r_m$. All policies share the same environment, but only one policy is updated per step. This means that the update of each policy is unaffected by the other policies, but only influenced indirectly through the changes in the robot’s state caused by other policies in the simulation environment.

For imitation learning–based multi-policy-transfer smoothness and robustness, the primary challenge is that out-of-domain disturbances during policy switching arise from large discontinuities in the reference motions encoded by different policies. To address this issue, we take a twofold approach: (i) introducing policy-transition scenarios during training, and (ii) injecting sudden perturbations to the robot’s body state through temporal discontinuities in the reference motions. In particular, to improve the smoothness and robustness against action transitions, we introduce two forms of randomization operation during training:

1) \textbf{Randomization for Policy Transition Robustness:} A key innovation in our training pipeline is the incorporation of a randomization method to explicitly train for transition scenarios, moving beyond naive single-skill imitation. During training, an episode does not necessarily consist of a single full motion playback. The length of a complete episode $T_{e}$ is defined as the sum of the durations of the all categories of reference motions $T_m,m\in M$. At any timestep $t$, with fixed probability $p_{trans}\in[0,1]$, a transition event is triggered stochastically as follows:
\[
b_t=\mathbf{Bernoulli}(p_{trans)}
\]
If $b_t=1$, the current policy $\pi^m$ and reference motion ${M}^r_m$ is abruptly truncated. Based on the current simulator and robot state, the update of the original policy network $\pi^m$ is frozen, and the training randomly switches to another action-specific policy $\pi^{m'}$, where $m'$ is sampled uniformly from the motion set $M$, with the reference motion $\mathcal{M}_t$ reset accordingly.
\[
\pi_{t+1}=\begin{cases}\pi^{m'},\mathrm{if~b_t=1}\\\pi^m,\mathrm{otherwise}\end{cases}
\]\\
\[
\mathcal{M}_{t+1}=\begin{cases}\mathcal{M}^r_{m',t=0},~\mathrm{if~b_t=1}\\ \mathcal{M}^r_{m,t=t+1},~\mathrm{otherwise}\end{cases}
\]
This process forces each expert $\pi^m$ to learn to start and terminate from a broad distribution of states, effectively simulating the situation that frequent interruptions and transitions in combat scenarios, and improving the robustness when policy switching.

2) \textbf{Randomization for Temporal Discontinuities:} When sampling a reference state from $\mathcal{M}^r_m$, we do not strictly follow consecutive frames. With a fixed probability $p_{jump}$, the reference motion at the next timestep $\mathcal{M}_{t+1}$ is randomly replaced by the reference motion from $k\sim \mathcal{U}\{1,K\}$ steps ahead.
\[
\mathcal{M}_{t+1}=\begin{cases}
\mathcal{M}^r_{m,t=t+k},~\mathrm{if~}\mathbf{Bernoulli}\mathrm{(p_{jump})=1}\\ \mathcal{M}^r_{m,t=t+1},~\mathrm{otherwise}
\end{cases}
\]
This compels the policy to handle non-smooth, discontinuous reference commands, significantly improving its robustness to timing misalignments during transitions.

\begin{algorithm}[t]
% \caption{Randomization Strategies for Robust Multi-Expert Policy Training}
\caption{Randomization Strategies}
\label{alg:randomization}
\begin{algorithmic}[1]
\STATE Load policies for all categories of motions
\STATE Choose one policy $\pi^m,m\in M$ to be activated, others are frozen
\FOR{each timestep $t$ in an episode}
    \STATE Sample current reference frame $\mathcal{M}_t=\mathcal{M}^r_{m, t}$
    \STATE Execute action $a_t \sim \pi^m(\mathcal{S}_t)$
    \STATE Update policy $\pi^m$

    \STATE {--- Randomization for Policy Transition ---}
    \IF{Bernoulli($p_{trans}$) = 1}
        \STATE Freeze update of $\pi^m$\\
        Sample New Reference Motion $m'\in M$
        \STATE Activate a new policy $\pi^m = \pi^{m'}$
        \STATE Reset reference motion $\mathcal{M}_{t+1}=\mathcal{M}^r_{m',t=0}$
    \ENDIF

    \STATE {--- Randomization for Temporal Discontinuities ---}
    \IF{Bernoulli($p_{jump}$) = 1}
        \STATE Replace $\mathcal{M}^r_{m,t+1}$ with $\mathcal{M}^r_{m,t+k}$, $k \sim \mathcal{U}(1,K)$
    \ENDIF\\
    Update States $\mathcal{S}_t$
\ENDFOR
\end{algorithmic}
\end{algorithm}

The aforementioned randomization techniques are formalized in pseudocode as presented in Algorithm \ref{alg:randomization}. This approach of imitation learning with randomized cross-policy and cross-temporal domain sampling effectively simulates two critical aspects of combat motions: the necessity to terminate ongoing actions and transition to new strategies at any moment, and the ability to handle abruptly changing motion references. Consequently, it significantly enhances the robustness of each individual expert policy.

% \subsection{Imitation Learning of Single Expert Policy.}
For the above expert networks, we employ an Asymmetric PPO algorithm for imitation learning training. All expert networks have an identical architecture. Upon completion of the described training process, we obtain multiple expert networks that can operate independently to execute their respective reference motions. In the Asymmetric PPO framework, the observations for the actor and critic networks are detailed in Table \ref{tab:observations}. All networks employ an MLP architecture. The output of all networks is a 23-dimensional joint actions $a\in\mathbb{R}^{23}$ for the robot.

\begin{table}[h]
\caption{Observation Space for the Expert Networks}
\label{tab:observations}
\centering
    \begin{tabularx}{1.0\linewidth}{lXX}
    \hline
    \textbf{Observation States} & \textbf{Actor Dims} & \textbf{Critic Dims} \\
    \hline
    Joint Positions & 23 & 23 \\
    Joint Velocities & 23 & 23 \\
    Root Angular Velocity & 3 & 3 \\
    Root Projected Gravity & 3 & 3 \\
    Actions & 23 & 23 \\
    Reference Motion Phase & 1 & 1 \\
    History (above) & $76\times4$ & $76\times4$ \\
    \hline
    Reference Joint Positions & 23 & 23 \\
    Reference Joint Velocities & 23 & 23 \\
    Reference Body Positions & 81 & 81 \\
    Future Reference Joint Positions & $23\times2$ & $23\times2$ \\
    Future Reference Body Positions & $81\times2$  & $81\times2$\\
    \hline
    Root Linear Velocity & - & 15 \\
    Body Position Difference & - & 81 \\
    Randomized Base CoM offset & - & 3 \\
    Randomized Base Link Mass & - & 22 \\
    \hline
    \textbf{Total Dimensions} & \textbf{715} & \textbf{836} \\
    \hline
    \end{tabularx}
\end{table}

We define many of the reward terms in the following format:

$$
\mathcal{R} = \sum_i w_i^eR(r_i)
$$

$$
R(r_i) = \exp{(-||r_i-r^{ref}_i||^2_2/\sigma_{r_i})}
$$
where the $w^e$ is weight for expert policies training, meanwhile $w^g$ firstly mentioned in Table. \ref{tab:rewards} is for the gating policy training. The update rule for the $\sigma_{s_i}$ value follows the same definition as presented in the PBHC~\cite{xie2025kungfubot}.

The reward function used during training is outlined in Table \ref{tab:rewards}. The reward is primarily divided into two components: one for motion tracking, and another for ensuring smooth task execution. For the latter component, we specifically introduce penalties on the absolute values of output torques and their temporal differences to ensure motion fluency. Furthermore, to prevent biologically implausible phenomena such as foot sliding during policy transitions, where feet might drag along the ground to reach target positions, we designed additional reward terms that penalize foot-ground sliding and encourage prolonged foot aerial phases. This design promotes lifting the feet during movement, better mimicking natural human motion characteristics.

\begin{table}[h]
\caption{Rewards of Expert Policies}
\label{tab:rewards}
\centering
\begin{tabularx}{1.0\linewidth}{@{}lccc@{}}
\hline
\textbf{Rewards} & \textbf{Expression} & \textbf{$w^e$} & \textbf{$w^g$} \\
\hline
Joint Positions & $R(q_t)$ & 1.0 & 0.3\\
Joint Velocities & $R(\dot q_t)$ & 1.0 & 0.3 \\
Body Positions & $R(p_t)$ & 2.0 & 0.6 \\
Body Rotations & $R(\theta_t)$ & 0.5 & 0.2 \\
Body Velocities & $R(\dot p_t)$ & 0.5 & 0.2 \\
Body Ang-Velocities & $R(\dot w_t)$ & 0.5 & 0.1 \\
Upper Body Positions & $R(p^{up}_t)$ & 4.0 & 0.2 \\
Feet Positions & $R(p^{feet}_t)$ & 1.0 & 0.6 \\
Max Joint Positions & $\exp{(-||q_t-q^r_t||_\infty/\sigma_{mj})}$ & 1.0 & 1.0 \\
\hline
Joint Position Limits & $\mathbb{I}(q_t\notin [q_{min}, q_{max}])$ & -10.0 & -5.0 \\
Joint Velocity Limits & $\mathbb{I}(\dot q_t\notin [\dot q_{min}, \dot q_{max}])$ & -5.0 & -3.0 \\
Joint Torque Limits & $\mathbb{I}(\tau_t\notin [\tau_{min}, \tau_{max}])$ & -5.0 & -4.0 \\
Feet Contact Forces & $\min(||F^{feet}-400||_2^2,0)$ & -1e-2 & -1e-2 \\
Feet Air Time & $\mathbb{I}[T_{air}>0.3]$ & -1.0 & -1.0 \\
Feet Slipping Penalty & $||v^{feet}||^2_2\cdot \mathbb{I}[||F^{feet}||_2>1]$ & -3.0 & -2.0 \\
Torque & $||\tau||^2_2$ & -1e-6 & -1e-6 \\
Collision Penalty & $\mathbb{I}_{collision}$ & -30.0 & -20.0 \\
Termination Penalty & $\mathbb{I}_{termination}$ & -300.0 & -300.0 \\
Alive & 1 & 0.8 & 0.8 \\
\hline
Torque Difference & $||\tau_t-\tau_{t-1}||^2_2$ & - & -0.5 \\
Task ID Check & $\exp(-||c_t-c_t^r||^2_2)$ & - & 2.0 \\
\hline
\end{tabularx}
\end{table}

During the training process, we also incorporated elements of curriculum learning by progressively tightening the precision requirements for motion tracking errors, thereby enhancing the convergence effectiveness of the policy.

\subsection{Gating Policy Training.}

Once the individual expert policies are obtained, and to involve the smoothness constraints on controlling, we freeze their parameters and design a gating network to produce smooth transitions while leveraging the motion capabilities of existing expert policies. Since the upper and lower limbs play different roles in the tracking task during motion execution, we define the dimension of the gating network output $\hat w$ as twice the number of expert policies $\hat w\in\mathbb{R}^{2N}$. Specifically, separate weight coefficients are assigned to the outputs of each expert policy for the upper and lower limbs, which are then applied independently to generate the final actions. The gating network receives the current robot states $\mathcal{S}_t$, which are the same as the input states of expert policies, and an optional task embedding $c_t$ that encodes the high-level combat instruction. It outputs a weight vector $\hat w_t = [\hat w_t^{u, 1}, \dots, \hat w_t^{u,N},\hat w_t^{l,1},\dots,\hat w_t^{l,N}]$ through a softmax layer, where $N$ denotes the number of expert policies, $\{u,l\}$ represents the weights for upper and lower body actions. The final control action $a^g$ is then computed as a weighted mixture of the experts:
$$
a^g_t = \sum_{p}\sum_m\hat w_t^{p,m} \pi^m_p(\mathcal{S}_t,c_t),\\p\in \{u, l\},m\in M
$$
specifically, $c_t$ is an $N$-dimensional one-hot vector, where the $i$-th entry is set to 1 for specific motion $\mathrm{index}(m)$, and all other entries are 0.

\[
c_{t,i}=\begin{cases}
    1,~\mathrm{if~i=index(m)}\\0,~\mathrm{otherwise}
\end{cases}
\]

The training procedure of the gating network follows the same scheme as that of the expert networks, adopting an asymmetric PPO structure. The actor policy is employed as a MLP network, followed by a softmax output. This lightweight architecture ensures real-time inference on physical hardware. In addition, we regularize the gating output with a temporal smoothness constraint to discourage rapid fluctuations between experts, which can otherwise result in discontinuous or unstable motions.

Its observation state is identical to that listed in Table \ref{tab:observations}, but further augmented with a task vector $c_t$ in actor network. During algorithm updates, the gating network places greater emphasis on task scheduling, as well as the smoothness and stability of output torques. Therefore, its reward design differs from that of the expert networks. As shown in Table \ref{tab:rewards}, the single-step reward is defined as:
$$
\mathcal{R} = \sum_i w_i^gR(r_i)
$$

\subsection{Validation and Deployment.}
After completing the training of all experts and gating policies, we validated our framework in the MuJoCo simulation environment and subsequently deployed it onto the Unitree G1 humanoid robot.

Additionally, we specifically designed a control pipeline for robotic combat tasks. 
In the real-world control pipeline, action commands are issued via a game joystick similar to playing an RPG action game: holding down a specific button triggers the execution of the corresponding action policy, while the robot defaults to locomotion mode when no command is given. 
For locomotion, we integrate a policy trained with the \texttt{RoboMimic} framework, ensuring stable operation of the robot outside combat tasks.
We implemented a complete control pipeline that enables RPG-style interaction: the robot can execute any fighting skill on demand, smoothly stop and switch between different actions at arbitrary timesteps, and seamlessly return to a locomotion state when no fighting commands are issued. 
\section{Experiments}

\subsection{Experimental Setup.}
We ultimately recorded monocular video data and converted them into SMPL-format motions, which were then retargeted to the Unitree G1 humanoid robot. Four representative fighting skills are considered: \textit{punching}, \textit{jumping}, \textit{sword swing}, and \textit{kicking}. Both the \textit{punching} and \textit{sword swing} motions are composed of multiple consecutive striking actions combined to form a complete movement. Based on the proposed RPG framework, we obtained multiple imitation-learning expert policies capable of executing individual actions, as well as a multi-policy control pipeline that enables stable and seamless policy transitions. The experiments were trained on the IsaacGym simulation platform using an RTX 4090 GPU, with sim-to-sim policy validation performed in MuJoCo prior to deployment. The control frequency on the real robot is 50 Hz. Finally, the proposed framework was successfully deployed on the real Unitree G1 robot. As illustrated in Fig. \ref{fig:motion} and Fig. \ref{fig:recover}, the final demonstration shows that the robot not only executes individual expert actions effectively, but also achieves smooth and stable transitions across multiple action policies.

The proposed RPG framework makes two primary contributions: (i) a training methodology incorporating policy-transition and temporal randomization mechanisms, and (ii) a gating network for policies merging. To validate the effectiveness of the proposed RPG framework, we conduct ablation studies, which will be discussed in the following sections.

\subsection{Baseline and Metrics.}
We employ the ASAP framework as the baseline for experimental comparisons, and all subsequent comparative experiments are conducted based on variants of the ASAP framework.

To evaluate the effectiveness of imitation learning for motion tracking, we introduce the following metrics to evaluate joint tracking performance: $E_{mpjpe}$(mean per-joint position error), $E_{mpjae}$(mean per-joint angle error), $E_{mpjve}$(mean per-joint velocity error), and metrics to assess root body tracking: $E_{rootpe}$(root position error), $E_{rootre}$(root rotation error), $E_{rootve}$(root velocity error).

To evaluate the robustness and control stability of the algorithm during cross-policy transitions, we propose a success rate metric denoted as $E_{succ,n}^{a\rightarrow b}$, which represents the success rate of the robot completing the transition from motion $a$ to motion $b$ across $n$ experimental trials. The motion indexes can be described as $j$ for jumping, $p$ for punching, $s$ for sword swing, and $k$ for kicking.

Additionally, we introduce $E_{maccj}$ (mean acceleration of joints) to evaluate the smoothness of robot control.

\subsection{Main Results.}
For the overall framework, our investigation focuses on the following key questions:

\textbf{Q1: The impact of policy-transition and temporal randomization on individual expert networks.}

To explore the impact of the RPG method on individual expert policies, we design a set of controlled experiments. The baseline approach involves using the ASAP framework to train imitation learning policies for each action separately. In contrast, the experimental group employs the RPG framework, where multiple expert networks are trained simultaneously using policy-transition and temporal randomization. After the training convergence, the corresponding individually trained expert networks from RPG are compared with those from the baseline.

\begin{table}[ht]
    \centering
    \caption{Expert Policies Comparison of Tracking Effectiveness}
    \small
    \label{table:tracking_effectiveness}
    \resizebox{\linewidth}{!}{%
    \begin{tabular}{l|l|c c c c c c}
        \toprule
        \multirow{2}{*}{\textbf{Motions}} & \multirow{2}{*}{\textbf{Experiments}} & \multicolumn{6}{c}{\textbf{Metrics}} \\
        \cmidrule(lr){3-8}
        & & $E_{mpjpe}$ & $E_{mpjae}$ & $E_{mpjve}$ & $E_{rootpe}$ & $E_{rootre}$ & $E_{rootve}$ \\
        \midrule
        \multirow{2}{*}{Jumping} 
        & baseline & \textbf{0.1492} & \textbf{0.0346} & \textbf{1.6895} & 0.1521 & 1.5218 & \textbf{0.2915} \\
        & RPG (ours) & 0.1594 & 0.0382 & 1.8231 & \textbf{0.1387} & \textbf{1.3874} & 0.3342 \\
        \midrule
        \multirow{2}{*}{Punching} 
        & baseline & 0.1428 & 0.0315 & \textbf{1.5218} & 0.1289 & \textbf{1.2186} & \textbf{0.2489} \\
        & RPG (ours) & \textbf{0.1321} & \textbf{0.0283} & 1.6254 & \textbf{0.1164} & 1.3247 & 0.2753 \\
        \midrule
        \multirow{2}{*}{Sword Swing} 
        & baseline & 0.1689 & \textbf{0.0321} & 1.8927 & \textbf{0.1298} & 1.4672 & 0.3028 \\
        & RPG (ours) & \textbf{0.1524} & 0.0368 & \textbf{1.7346} & 0.1423 & \textbf{1.3429} & \textbf{0.2784} \\
        \midrule
        \multirow{2}{*}{Kicking} 
        & baseline & \textbf{0.1412} & \textbf{0.0302} & 1.7825 & 0.1357 & 1.3984 & \textbf{0.2631} \\
        & RPG (ours) & 0.1537 & 0.0339 & \textbf{1.6428} & \textbf{0.1225} & \textbf{1.2873} & 0.2896 \\
        \bottomrule
    \end{tabular}%
    }
\end{table}

We evaluated the policies trained for jumping, punching, sword swing, and kicking using the aforementioned motion tracking metrics. For lower-body dominant motions such as jumping and kicking, the tracking performance of RPG is generally weaker. This may be attributed to the greater disturbance impact on the lower limbs when handling abrupt motion transitions. Meanwhile, the baseline method shows relatively stronger performance in velocity tracking, likely because the robot’s control is subject to safety constraints when sudden changes in reference motion cause sharp velocity variations.

However as the results, as shown in the table, indicate that there is no significant difference between the two methods in terms of motion tracking performance.

\textbf{Q2: The effect of the proposed framework on robustness in multi-policy transitions.}

To evaluate the robustness of the control methods, we aim to determine which approach enables the robot to maintain better motion completion when facing out-of-domain states and disturbances during policy transitions. For this purpose, we employ the $E_{succ,n}$ metric to assess whether the robot can successfully execute motion switches without falling. In this set of baseline experiments, expert networks trained using the ASAP framework are directly integrated into the pipeline to perform policy switching. For the experimental group, the pipeline incorporates the framework trained with the RPG method. Each motion transition is repeated 20 times, and the success rate is calculated.

\begin{table}[ht]
    \centering
    \caption{Success Rate of Transition}
    \small
    \label{table:success_rate_transition}
    \resizebox{\linewidth}{!}{%
    \begin{tabular}{l|l|c c c c}
        \toprule
        \multirow{2}{*}{\textbf{Start (a)}} & \multirow{2}{*}{$E_{succ,20}^{a\rightarrow b}$} & \multicolumn{4}{c}{\textbf{Target Motion (b)}} \\
        \cmidrule(lr){3-6}
        & & Jumping & Punching & Sword Swing & Kicking \\
        \midrule
        \multirow{2}{*}{Jumping} 
        & baseline & 0.25 & 0.15 & 0.15 & 0.05 \\
        & RPG (ours) & \textbf{0.70} & \textbf{0.85} & \textbf{0.80} & \textbf{0.75} \\
        \midrule
        \multirow{2}{*}{Punching} 
        & baseline & 0.35 & 0.60 & 0.50 & 0.45 \\
        & RPG (ours) & \textbf{0.80} & \textbf{0.95} & \textbf{0.75} & \textbf{0.75} \\
        \midrule
        \multirow{2}{*}{Sword Swing} 
        & baseline & 0.30 & 0.50 & 0.45 & 0.65 \\
        & RPG (ours) & \textbf{0.70} & \textbf{0.95} & \textbf{0.90} & \textbf{0.90} \\
        \midrule
        \multirow{2}{*}{Kicking} 
        & baseline & 0.30 & 0.65 & 0.40 & 0.50 \\
        & RPG (ours) & \textbf{0.70} & \textbf{0.90} & \textbf{0.75} & \textbf{0.85} \\
        \bottomrule
    \end{tabular}%
    }
\end{table}

As shown in the results Table. \ref{table:success_rate_transition}, the success rate of motion transitions is closely related to the primary body parts involved in the movements. For example, transitions between lower-body motions—such as from jumping to kicking or from jumping to jumping—exhibit the highest level of difficulty. It also shows that RPG framework significantly improves the success rate of motion transitions. This demonstrates that the RPG framework exhibits stronger robustness in scenarios involving multi-policy switching.

\textbf{Q3: The effect of the proposed framework on control smoothness in multi-policy transitions.}

Ensuring smooth motion execution is critically important when handling abrupt action changes or cross-policy scenarios. Our experimental design follows the same setup as described for Q2, and we select $E_{maccj}$ as the metric for evaluation. To assess control smoothness during policy transitions, we analyze a time window spanning from 50 timesteps before to 50 timesteps after each motion transition.

\begin{table}[ht]
    \centering
    \caption{Smoothness Comparison while Transition}
    \small
    \label{table:smooth_compare}
    \resizebox{\linewidth}{!}{%
    \begin{tabular}{l|l|c c c c}
        \toprule
        \multirow{2}{*}{\textbf{Start (a)}} & \multirow{2}{*}{$E_{maccj}$} & \multicolumn{4}{c}{\textbf{Target Motion (b)}} \\
        \cmidrule(lr){3-6}
        & & Jumping & Punching & Sword Swing & Kicking \\
        \midrule
        \multirow{3}{*}{Jumping} 
        & baseline & 1.8520 & 1.9533 & 1.8925 & 1.7633 \\
        & RPG w/o Gating & 1.9197 & 1.8649 & 1.9024 & 1.5323 \\
        & RPG (ours) & \textbf{1.5235} & \textbf{1.6421} & \textbf{1.5875} & \textbf{1.4524} \\
        \midrule
        \multirow{3}{*}{Punching} 
        & baseline & 1.9233 & \textbf{1.2383} & 1.8738 & 1.9642 \\
        & RPG w/o Gating & 1.8125 & 1.3096 & 1.9482 & 1.9773 \\
        & RPG (ours) & \textbf{1.6237} & 1.3529 & \textbf{1.7821} & \textbf{1.6535} \\
        \midrule
        \multirow{3}{*}{Sword Swing} 
        & baseline & 1.8735 & \textbf{1.8127} & 1.3525 & 1.8239 \\
        & RPG w/o Gating & 1.7769 & 1.8322 & 1.2617 & 1.8254 \\
        & RPG (ours) & \textbf{1.5841} & 1.9031 & \textbf{1.1424} & \textbf{1.5432} \\
        \midrule
        \multirow{3}{*}{Kicking} 
        & baseline & 1.9637 & 1.9826 & 1.8936 & 1.7235 \\
        & RPG w/o Gating & 1.7689 & 1.8236 & 1.9341 & 1.8099 \\
        & RPG (ours) & \textbf{1.6539} & \textbf{1.6724} & \textbf{1.5937} & \textbf{1.4528} \\
        \bottomrule
    \end{tabular}%
    }
\end{table}

As shown in the Table. \ref{table:smooth_compare}, the smoothness of motion is highly correlated with the range of motion. For instance, transitions involving punching can achieve relatively smooth results even with the baseline method. For motions with large amplitude action such as jumping and sword-swinging, the RPG generates significantly smoother motion executions. Even in cases where the smoothness of motions generated by RPG is inferior to that of the baseline, the difference between the two is not significant. The results indicate that the RPG method yields significantly smoother control outputs across the majority of motion transitions.

\textbf{Q4: How does the gating policy perform in merging the expert policies?}

\begin{figure}[t]
  \centering
  \includegraphics[width=\columnwidth]{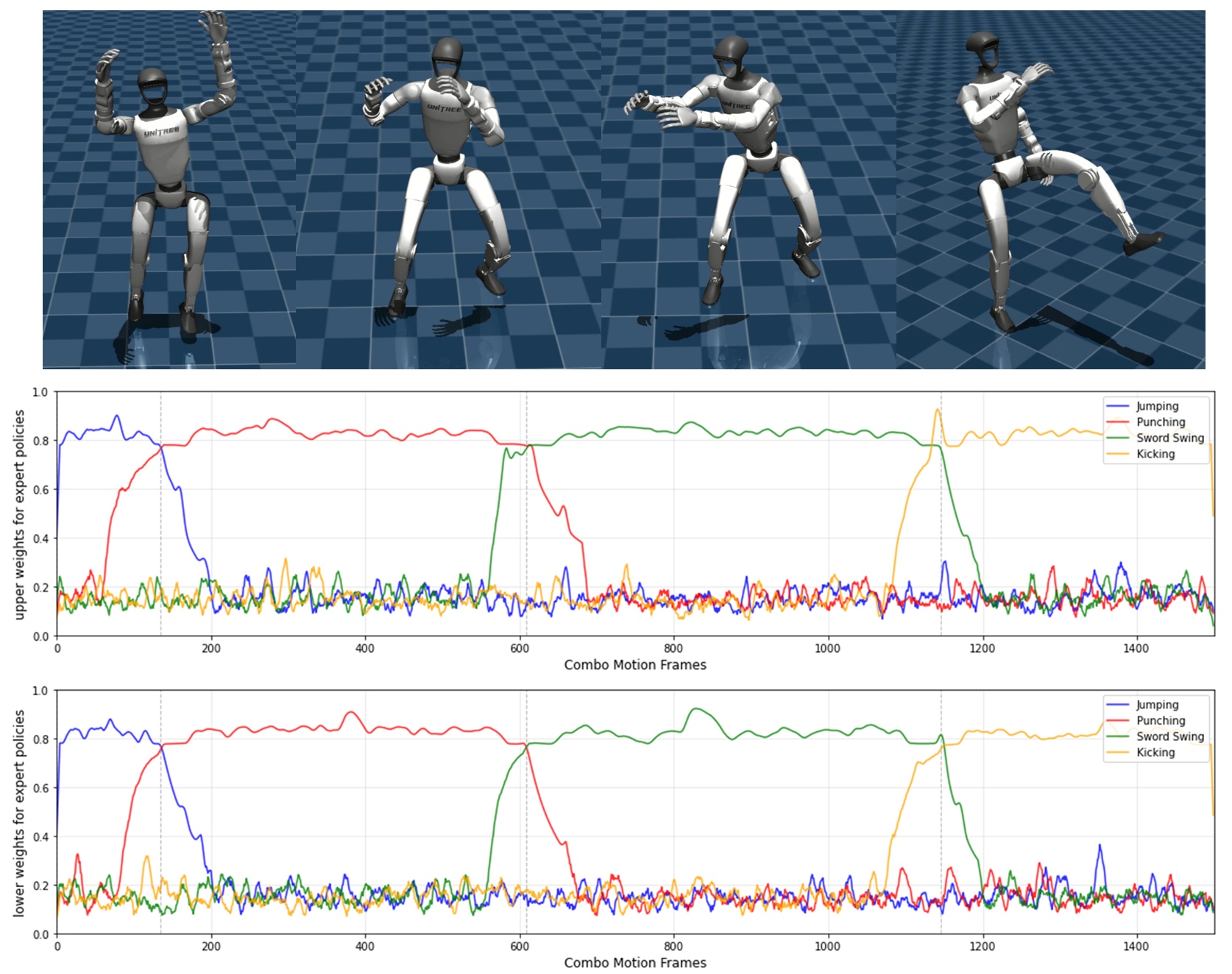}
  \caption{ Policy Transition Example: Jumping-Punching-Sword Swing-Kicking. The upper curve graph represents the variation of $\hat w^{u,m}_t$ along the time. The lower curve graph represents the $\hat w^{l,m}_t$ variation.
  }
  \label{fig:gating_output}
\end{figure}

We selected a motion transition combo as an example and recorded the gating policy output. The results are shown in Fig. \ref{fig:gating_output}. Since the weight values output by the gating policy account for the coordination between the upper and lower body of the robot, we recorded them separately. It can be observed that during action transitions, the weight values corresponding to the executed actions output by the gating network are more pronounced.

We also designed an ablation experiment comparing the performance with/without the gating mechanism. As shown in the Table. \ref{table:smooth_compare}, it tells that the gating policy significantly improves the control smoothness during policy transitions.

\subsection{Discussion.}

Based on the aforementioned experiments, it can be observed that the use of the RPG significantly enhances the stability and success rate of action policy transitions while improving the smoothness of robot control—all without substantially compromising the motion tracking performance of individual expert policies. This improvement can be attributed, in part, to the policy-transition and temporal randomization mechanisms, which expand the robot's state domain, and through RL reward constraints, enhance both control smoothness and disturbance resistance in the face of abrupt motion changes.

% \begin{figure*}[t]
%   \centering
%   \includegraphics[width=\columnwidth]{figures/rpg_weight_curve.jpg}
%   \caption{\textbf{Gating Network Outputs in Combo.} Collection of human demonstration }
%   \label{fig:gating_output}
% \end{figure*}

\section{CONCLUSION}

This paper proposed RPG, a robust policy gating framework for multi-policies transition in humanoid fighting. RPG integrates policy-transition and temporal randomization mechanism for training with a smoothly regularized gating network, enabling stable and interruptible motion execution. Simulations and real-world tests on a Unitree G1 robot demonstrate that RPG achieves high transition success rates and improved control smoothness and robustness without compromising tracking accuracy. Meanwhile we developed a multi-policy control pipeline that enables robot combat tasks to be executed in a manner reminiscent of controlling a character in an action RPG game. Our approach provides a practical solution for dynamic multi-skill imitation in real-world humanoid applications.

In future work, we plan to integrate perception and recognition capabilities to enable the robot to autonomously track and engage targets in combat tasks.

% \newpage
\bibliography{root}

\end{document}